\documentclass[12pt]{article}
\usepackage[T1]{fontenc}
\usepackage[english]{babel}
\usepackage{amsmath,amssymb,amsthm}
\usepackage{booktabs}       % professional-quality tables
\usepackage{amscd}
\usepackage{mathrsfs} %pour mathscr
\usepackage[shortlabels]{enumitem}
\usepackage{euscript}
\usepackage{graphicx}
\usepackage{amscd,bm}
\RequirePackage[colorlinks,citecolor=blue,urlcolor=blue]{hyperref}

\newtheorem{theorem}{Theorem}[section]
\newtheorem{corollary}[theorem]{Corollary}

\newtheorem{proposition}[theorem]{Proposition}

\theoremstyle{definition}

\newtheorem{algorithm}[theorem]{Algorithm}
\newtheorem{remark}[theorem]{Remark}
\newtheorem{example}[theorem]{Example}

\newcommand{\abs}[1]{{\lvert {#1}\rvert}}
\newcommand{\norm}[1]{{\lVert {#1}\rVert}}

\newcommand{\scal}[2]{{\left\langle{{#1},{#2}}\right\rangle}}

\DeclareMathOperator{\dom}{dom}
\DeclareMathOperator{\sign}{sign}

\newcommand{\XC}{\mathcal{X}}

\newcommand{\YC}{\mathcal{Y}}

\newcommand{\prox}{\mathrm{prox}}

\newcommand{\R}{\mathbb{R}}

\newcommand{\KK}{\mathbb{K}}

\newcommand{\N}{\mathbb{N}}

\begin{document}

\title{ {\bf 
Solving $\ell^p$-norm regularization with tensor kernels
}} 

\author{Saverio Salzo$^1$ and Johan A.K. Suykens$^2$ and Lorenzo Rosasco$^{1,3}$\\[3mm]
\small
\small $\!^1$LCSL, Istituto Italiano di Tecnologia and
Massachusetts Institute of Technology\\
\small Via Morego 30, 16163 Genova, Italy\\
%\small Bldg.~46-5155, 77 Massachusetts Avenue, Cambridge, MA 02139, USA\\
\small \texttt{saverio.salzo@iit.it}\\[3mm]
\small
\small $\!^2$KU Leuven, ESAT-STADIUS\\
\small Kasteelpark Arenberg 10, B-3001 Leuven (Heverlee), Belgium \\
\small \texttt{johan.suykens@esat.kuleuven.be}\\[3mm]
\small
\small $\!^3$DIBRIS, Universit\`a degli Studi di Genova\\
\small Via Dodecaneso 35, 16146 Genova, Italy \\
\small \texttt{lrosasco@mit.edu}
}

\date{}
\maketitle
{\abstract 
In this paper, we discuss how a suitable family of tensor kernels can be used to efficiently solve  
nonparametric extensions of $\ell^p$ regularized learning methods. 
Our main contribution is proposing a fast dual algorithm, and showing that it allows 
to solve the problem efficiently. Our results contrast recent findings suggesting kernel methods 
cannot be extended beyond Hilbert setting. 
Numerical experiments confirm the effectiveness of the method.
}

\section{Introduction}

Kernel methods are classically formulated as a regularized empirical risk 
minimization and yields flexible and effective non-parametric models. However, they are restricted to 
$\ell^2$-regularization. Indeed the so called \emph{kernel trick} crucially rely on 
a scalar product structure (a Hilbert space). The basic tool
of these methods is the kernel function which, evaluated at the training points, 
allows $(a)$
 to formulate a ``dual'' optimization problem, which is essentially quadratic and finite dimensional,
and $(b)$, 
through the solution of dual problem, to obtain  an explicit 
linear representation of the solution of the original (primal) problem
(\emph{the representer theorem}) \cite{Ste2008,Vap98}. 
This dual approach provides a feasible way to deal with 
non-parametric (infinite dimensional) models, and
a possibly easier and more efficient algorithm to tackle
 the finite dimensional also.

It is well known that kernels for other norms can be defined 
\cite{Song13,Zhan09,Zhan12}, but recent results suggest that
they are unpractical \cite{nips11}. In particular, 
these kernels do not allows to properly express, in closed-form, the dual problem,
making the kernel trick inapplicable.
In this paper, we question this conclusion.
We consider $\ell^p$-regularization for $1<p<2$ and starting from \cite{sal16b}
we illustrate how, for certain values of $p$, a class of tensor kernels make it 
possible to derive a dual problem that can be efficiently solved.
Our main contribution is a dual algorithm, having fast convergence properties,
that provides a way to overcome the well-known computational issues related
to non-Hilbertian norms, and makes the kernel trick 
still viable. From the optimization point of view, the challenge is that
some standard assumptions are not satisfied. Indeed the
dual objective function lacks a global Lipschitz continuous gradient, since
it incorporates a convex polynomial of degree strictly greater than 2.
Moreover, depending on the choice of the loss, constraints may be present.
Considering all these aspects,
the proposed algorithm is a dual proximal gradient 
method with linesearch which in the case of the least square loss and logistic loss 
we prove to converge linearly.
Numerical examples show the effectiveness of the proposed
framework and the possible application for variable selection.

The rest of the paper is organized as follows.
In section~\ref{sec:main} we explain how tensor kernels arise in $\ell^p$ regularization 
learning problems 
and provide an efficient algorithm to solve such problems, which is the main contribution of the paper.
In section~\ref{sec:theory} one finds the main elements of the theoretical analysis.
Finally, section~\ref{sec:numerics} contains the numerical experiments.

\paragraph{Notation.}
If $p>1$, $q>1$ is its conjugate exponent, i.e. 
$1/p + 1/q = 1$.  
Vectors are denoted by bold fonts and scalars by plain fonts. For every 
$\bm{x},\bm{x}^\prime \in \R^d$, 
 $\bm{x}\odot\bm{x}^\prime \in \R^d$  and 
 $\bm{x}\otimes\bm{x}^\prime \in \R^{d\times d}$  
are their Hadamard and tensor product respectively, and 
$\mathrm{sum}(\bm{x})\in \R$ denotes the sum of the components of $\bm{x}$.
If $\KK$ is a countable  set,  we denote by $\ell^p(\KK)$ the space of $p$-summable sequences indexed in $\KK$
 with $p$-norm $\norm{\bm{w}}_p= \big(\sum_{k \in \KK} \abs{w_k}^p\big)^{1/p}$.
We define the \emph{duality map} of $\ell^q(\KK)$ as $J_q \colon \ell^q(\KK) \to \ell^p(\KK)$
with $J_q(\bm{u}) = (\sign(u_k) \abs{u_k}^{q-1})_{k \in \KK}$ \cite{Sch2012}. 

\section{Motivation and main contribution}
\label{sec:main}

First, we recall how kernel methods arise for $\ell^2$-regularization.
Next, we present the objective of this study, i.e.,
an effective $\ell^p$-norm regularized learning method.
Based on \cite{sal16b}, which showed
 that this method
can be \emph{kernelized} by an appropriate \emph{tensor kernel},
we present a novel dual algorithm which uses the knowledge of the tensor kernel only
and converges linearly.

\subsection{Classical kernel methods}
\label{subsec:back}
We begin with a look at a simple kernel method, that is, \emph{kernel ridge regression},
and we highlight the role played by duality.
Later, this will serve as a guide to generalize the theory to $\ell^p$-regularization. 
Ridge regression is formulated as the following optimization problem
\begin{equation}
\label{eq:ridge}
\min_{\bm{w} \in \R^d} \frac{\gamma}{2} \norm{\bm{X} \bm{w} -\bm{y}}_2^2 + \frac 1 2 \norm{\bm{w}}_2^2,
\end{equation}
where $\bm{X} \in \R^{n\times d}$ is the data matrix.
This problem has a companion dual problem which is
\begin{equation}
\label{eq:dualridge}
\min_{\bm{\alpha} \in \R^n} \frac 1 2 \norm{\bm{X}^* \bm{\alpha}}_2^2 
+ \frac{1}{2 \gamma} \norm{\bm{\alpha}}_2^2 - \scal{\bm{y}}{ \bm{\alpha}}.
\end{equation}
These two problems are indeed related: writing the optimality conditions for 
\eqref{eq:ridge} and \eqref{eq:dualridge}
one obtains 
\begin{equation*}
\bm{X}^*(\bm{X} \bm{w} - \bm{y}) + \gamma^{-1} \bm{w}=0\qquad\text{and}
\qquad\bm{X}\bm{X}^*\bm{\alpha} - \bm{y} + \gamma^{-1} \bm{\alpha} = 0
\end{equation*}
respectively; and hence it immediately follows that if $\bar{\bm{\alpha}}$ is the solution of \eqref{eq:dualridge}, then
\begin{equation}
\label{eq:representer}
\bar{\bm{w}} = \bm{X}^* \bar{\bm{\alpha}} = \sum_{i=1}^n \bar{\alpha}_i \bm{x}_i
\end{equation}
is the unique solution of \eqref{eq:ridge}. Equation \eqref{eq:representer} is the content of the so called 
\emph{representer theorem} which ensures that the solution of 
a regularized regression problem can be written as a linear combination of the 
data points $\bm{x}_i \in \R^d$, $i = 1, \cdots, n$. 
Moreover, for the linear estimator it holds
\begin{equation}
\label{eq:representer2}
\scal{\bar{\bm{w}}}{ \bm{x}} = \sum_{i=1}^n \bar{\alpha}_i \scal{\bm{x}_i}{ \bm{x}} 
= \sum_{i=1}^n \bar{\alpha}_i K(\bm{x}_i,\bm{x}),
\end{equation}
where $K\colon \R^d \times \R^d \to \R$ is the \emph{linear kernel function} defined as
$K(\bm{x},\bm{x}^\prime) = \scal{\bm{x}}{\bm{x}^\prime}$. We note that, since
$\bm{X} \bm{X}^* = (K(\bm{x}_i,\bm{x}_j))_{1 \leq i \leq n, 1 \leq j \leq n}$, 
the dual problem \eqref{eq:dualridge} can also be written in terms
of the linear kernel function, taking the form of the following quadratic optimization problem
\begin{equation}
\label{eq:dualridge2}
\min_{\bm{\alpha} \in \R^n} \frac 1 2 
\sum_{i,j=1}^n K(\bm{x}_i,\bm{x}_j) \alpha_i \alpha_j
+ \frac{1}{2\gamma} \scal{\bm{\alpha}}{\bm{\alpha}} - \scal{\bm{y}}{\bm{\alpha}}.
\end{equation}
So, summarizing, the dual problem \eqref{eq:dualridge2}
and the representation formulas \eqref{eq:representer}-\eqref{eq:representer2}  provide a way to solve
the primal problem \eqref{eq:ridge} and to evaluate 
the optimal linear estimator by relying on the knowledge of the linear kernel function only.
This conclusion can then be extended to nonlinear regression models, by introducing general
kernel functions defined as
\begin{equation}
\label{eq:kernel}
K(\bm{x},\bm{x}^\prime) = \scal{\Phi(\bm{x})}{\Phi(\bm{x}^\prime)} 
= \mathrm{sum} (\Phi(\bm{x}) \odot \Phi(\bm{x}^\prime)),
\end{equation}
for some nonlinear \emph{feature map} $\Phi\colon \R^d \to \ell^2$.
This is the so called \emph{kernel trick} and it is at the basis of \emph{kernel methods} in machine learning, allowing even to treat infinite dimensional (nonparametric) models.
Kernels, defined by \eqref{eq:kernel}, can indeed be characterized as 
\emph{positive definite functions}, in the sense that
for every $n \in \N$, $(\bm{x}_i)_{1 \leq i \leq n} \in {(\R^d)}^n$, and
$\bm{\alpha} \in \R^n$, $\sum_{i,j=1}^nK(\bm{x}_i,\bm{x}_j)\alpha_i \alpha_j \geq 0$.
Moreover, kernels define an associated function space which is a reproducing kernel Hilbert space. There are many significant examples of kernel functions
and we cite among the other the Gaussian kernel $K(\bm{x},\bm{x}^\prime) 
= \exp(- \eta^{-2} \norm{\bm{x} - \bm{x}^\prime}_2^2)$ and the polynomial 
kernel $K(\bm{x},\bm{x}^\prime) = \scal{\bm{x}}{\bm{x}^\prime}^s$, 
describing the space of homogeneous polynomials of degree $s$.
We note that the theory can be further generalized to handle more general loss functions,
so to include classification problems too \cite{Ste2008,Vap98}.

\subsection{Kernel methods beyond $\ell^2$-regularization}

In view of the discussion above, 
a natural question is whether kernel methods can be extended to other regularization terms.
In particular $\ell^1$-regularization would be important in view 
of its properties to provide sparse solutions. Unfortunately,
in general $\ell^1$-regularization methods cannot be \emph{kernelized} (although they admit dual) \cite{Has15,Osb00} and
a useful representer theorem and definition of kernel 
can be obtained only under severe restrictions \cite{Song13}.
However, it was noted in \cite{Kol09} that $\ell^p$-regularization can be seen as a proxy to $\ell^1$
for suitable $p$. Moreover, it was recently shown in \cite{sal16b} 
that for certain values of $p \in \left]1,2\right[$ (arbitrarily close to $1$), the 
 $\ell^p$-regularization method can indeed be \emph{kernelized}, 
 provided that a suitable definition of tensor kernel is introduced. 
Here we recall the theory in \cite{sal16b}
for a simple model in order to make it more transparent.
Thus, in analogy to section~\ref{subsec:back}, we consider the problem
\begin{equation}
\label{eq:ellp}
\min_{\bm{w} \in \R^d} \frac{\gamma}{2} \norm{\bm{X} \bm{w} - \bm{y}}_2^2 
+ \frac 1 p \norm{\bm{w}}_p^p:=F(\bm{w}),
\end{equation}
where $1<p<2$.
In this case the dual problem is
\begin{equation}
\label{eq:dualellp}
\min_{\bm{\alpha} \in \R^d}\frac{1}{q} \norm{\bm{X}^* \bm{\alpha}}_q^q 
+ \frac{1}{2 \gamma} \norm{\bm{\alpha}}_2^2 - \scal{\bm{y}}{\bm{\alpha}}:=\Lambda(\bm{\alpha}),
\end{equation}
where $q$ is the conjugate exponent of $p$ (that is $1/p + 1/q = 1$).
Now, following the same argument as in section~\ref{subsec:back}, we write the optimality conditions
of the two problems. Then we have
\begin{equation}
\label{eq:kktellp}
\bm{X}^*(\bm{X} \bm{w} - \bm{y}) + \gamma^{-1} J_p(\bm{w}) = 0\quad\text{ and }\quad
\bm{X} J_q(\bm{X}^*\bm{\alpha}) - \bm{y} + \gamma^{-1} \bm{\alpha} = 0,
\end{equation}
where $J_p \colon \R^d \to \R^d$ and $J_q\colon \R^d \to \R^d$ 
are the gradients of $(1/p) \norm{\cdot}_p^p$ and $(1/q) \norm{\cdot}_q^q$
respectively (they are the duality maps).
Thus, multiplying by $\bm{X}^*$ the second equation in \eqref{eq:kktellp} and taking into account that $J_p \circ J_q = \mathrm{Id}$, it follows that if 
$\bar{\bm{\alpha}}$ is the solution of \eqref{eq:dualellp}, then 
$\bar{\bm{w}} = J_q(\bm{X}^* \bar{\bm{\alpha}})$
is the solution of \eqref{eq:ellp}. So, in this case the \emph{representer theorem} becomes
\begin{equation}
\label{eq:20170511a}
\bar{\bm{w}} = J_q(\bm{X}^* \bar{\bm{\alpha}}) = J_q \bigg( \sum_{i=1}^n \bar{\alpha}_i \bm{x}_i\bigg).
\end{equation}
We remark that, in contrast to the $\ell^2$ case, the above representation 
 is nonlinear in the $\alpha_i$'s,
because of the presence of the nonlinear map $J_q$. Indeed this map acts component-wise as the derivative of $(1/q)\abs{\cdot}^q$, i.e.,  $\sign(\cdot) \abs{\cdot}^{q-1}$.
Therefore, at first sight it is not clear how to define 
an appropriate kernel function that can represent 
the estimator $\scal{\bar{\bm{w}}}{\bm{x}}$ in analogy to \eqref{eq:representer2}, and make 
the kernel trick still successful.
So, it comes as a surprise
 that this is possible if one makes the following assumption \cite{sal16b}
 \begin{equation}
 \label{q-hypothesis}
\boxed{
q  \text{ is an even integer and } q \geq 2.
}
\end{equation}
Indeed in that case, for every $\bm{u} \in \R^d$,
$J_q(\bm{u}) = 
\big(\sign(u_j) \abs{u_j}^{q-1}\big)_{1 \leq j \leq d}
=(u_j^{q-1})_{1 \leq j \leq d}$,
and hence, using \eqref{eq:20170511a}, we have
\begin{equation}
\label{eq:20170514a}
\scal{\bar{\bm{w}}}{\bm{x}} 
= \sum_{j=1}^d 
\bigg(\sum_{i=1}^n \bar{\alpha}_i x_{i,j}\bigg)^{q-1}\!\!\!\! x_j
= \sum_{j=1}^d \sum_{i_1,\dots, i_{q-1}=1}^n \!\!
x_{i_1,j}\cdots x_{i_{q-1},j} x_j \bar{\alpha}_{i_1} \cdots \bar{\alpha}_{i_{q-1}},
\end{equation}
where we could expand the power of the summation in a multilinear form since $q$ is an integer.
Therefore, we are defining the \emph{linear tensor kernel function} $K$ as
\begin{equation}
\label{eq:linearker}
K\colon \underbrace{\R^d \times \cdots \times \R^d}_{q\ \text{times}} \to \R,
\quad K(\bm{x}^\prime_{1}, \cdots, \bm{x}^\prime_q) =
\sum_{j=1}^d x^\prime_{1,j}\cdots x^\prime_{q,j} 
= \mathrm{sum} (\bm{x}^\prime_1 \odot \cdots \odot \bm{x}^\prime_q),
\end{equation}
so that, \eqref{eq:20170514a} turns to
\begin{equation}
\label{eq:20170511b}
\scal{\bar{\bm{w}}}{\bm{x}} =
\hspace{-2ex}
\sum_{i_1,\dots, i_{q-1}=1}^n  
\hspace{-2ex}K(\bm{x}_{i_1}, \cdots, \bm{x}_{i_{q-1}}, \bm{x} )
\bar{\alpha}_{i_1} \cdots \bar{\alpha}_{i_{q-1}}.
\end{equation}
Comparing \eqref{eq:linearker} and \eqref{eq:kernel} we recognize that we may interpret the tensor kernel 
as a kind of group-wise similarity measure in the input space.
Moreover, since $q$ is even,
\begin{equation*}
\norm{X^* \bm{\alpha}}_q^q = \sum_{j=1}^d \bigg(\sum_{i=1}^n \alpha_i x_{i,j}\bigg)^q
 =\sum_{j=1}^d 
\sum_{i_1, \dots, i_q = 1}^n  x_{{i_1},j} \cdots x_{{i_q},j} \alpha_{i_1}\cdots \alpha_{i_q}
\end{equation*}
and hence, by exchanging the two summations above, 
the dual problem \eqref{eq:dualellp} becomes
\begin{equation}
\label{eq:dualellp2}
\min_{\bm{\alpha} \in \R^d} \frac 1 q \sum_{i_1, \dots, i_q = 1}^n K(\bm{x}_{i_1}, \dots, \bm{x}_{i_q}) \alpha_{i_1}\cdots \alpha_{i_q}
+ \frac{1}{2 \gamma} \norm{\bm{\alpha}}^2 - \scal{\bm{y}}{\bm{\alpha}}.
\end{equation}
We see now that, instead of the quadratic problem \eqref{eq:dualridge2} we have a
convex polynomial optimization problem of degree $q$.\footnote{The problem is convex since
the first term in \eqref{eq:dualellp2} is equal to $(1/q)\norm{X^* \bm{\alpha}}_q^q$.} 
The introduction of the tensor kernel \eqref{eq:linearker} allows to parallel the $\ell^2$ case, in the sense that the dual problem \eqref{eq:dualellp2} and formula 
\eqref{eq:20170511b} 
 provide the solution of the
regression problem \eqref{eq:ellp}.
Once again, the method can be extended to general feature maps 
 $\Phi\colon \R^d \to \ell^q(\KK)$, $\Phi(\bm{x}) = (\phi_k(\bm{x}))_{k \in \KK}$, with $\KK$ a countable set, provided that, in the definition of $K$, $\bm{x}_i$ is replaced by $\Phi(\bm{x}_i)$. Thus, 
a general \emph{tensor kernel} is defined as 
\begin{equation}
\label{def:tensorkernel2}
K(\bm{x}^\prime_{1}, \cdots, \bm{x}^\prime_q) 
= \sum_{k \in \KK} \phi_k(\bm{x}^\prime_{1})\cdots \phi_k(\bm{x}^\prime_{q}) 
= \mathrm{sum} (\Phi(\bm{x}^\prime_1) \odot \cdots \odot \Phi(\bm{x}^\prime_q)).
\end{equation}
It is easy to show that tensor kernels are 
still symmetric and positive definite, in the sense that
\begin{itemize}[-]
\item 
$\forall\, \bm{x}^\prime_1, \dots, \bm{x}^\prime_q \in \R^d$, 
and every permutation $\sigma$ of  $\{1,\dots, q\}$, 
$K({\bm{x}}^\prime_{\sigma(1)} \dots {\bm{x}}^\prime_{\sigma(q)}) 
= K({\bm{x}}^\prime_{1}, \dots {\bm{x}}^\prime_{q})$;
\item for every
$\bm{x}^\prime_1, \dots, \bm{x}^\prime_n \in \R^d$ and every $\bm{\alpha} \in \R^n$,
$\sum_{i_1,\dots, i_q = 1}^n K({\bm{x}}^\prime_{i_1}, \dots, {\bm{x}}^\prime_{i_q}) \alpha_{i_1} \dots \alpha_{i_q} \geq 0$.\footnote{
However, it is not known whether a function 
$K\colon \big(\R^d \big)^q \to \R$ satisfying the two properties above can be written as in \eqref{def:tensorkernel2} for some feature map $\Phi\colon \R^d \to \ell^q(\KK)$.}
\end{itemize}

These tensor kernels define an associated function space 
which is now  a 
reproducing kernel Banach space (See Section A.3 in the supplementary material and \cite{sal16b,Zhan09}).
Moreover, reasoning as in \eqref{eq:20170514a}, the following representation formula
can be proved
\begin{equation}
\label{eq:20171009b}
\scal{\bar{\bm{w}}}{\Phi(\bm{x})} =\hspace{-2ex}\sum_{i_1,\dots, i_{q-1}=1}^n  
\hspace{-3ex}
K(\bm{x}_{i_1}, \cdots, \bm{x}_{i_{q-1}}, \bm{x} )
\bar{\alpha}_{i_1} \cdots \bar{\alpha}_{i_{q-1}}.
\end{equation}
Finally, there do exist cases in which tensor kernel functions
can be computed without knowing the feature map $\Phi$ itself. The 
following \emph{polynomial} and \emph{exponential} tensor kernels
 are examples of such cases (but, there are others in the class of
power series tensor kernels \cite{sal16b}). 

\paragraph{Polynomial tensor kernel of degree $s \in \N$, $s \geq 1:$}
\begin{equation*}
K(\bm{x}^\prime_1,\dots, \bm{x}^\prime_q) 
=   \Big(\sum_{j=1}^d x^\prime_{1,j}  \cdots x^\prime_{q,j} \Big)^{s}
= \big(\mathrm{sum}(\bm{x}^\prime_1 \odot \cdots \odot \bm{x}^\prime_q) \big)^s.
\end{equation*}
It describes the space of homogeneous polynomials in $d$ real variables of degree $s$.
This corresponds to a finite dimensional model for which 
$\KK = \big\{ k \in \N^d \,\big\vert\, \sum_{j=1}^d k_j = s\big\}$ and, for every $k \in \N^d$,
$\phi_k(\bm{x}) = \big( s!/(k_1! \cdots k_d !)\big)^{1/q} \bm{x}^k$,
that is $(\phi_k)_{k \in \KK}$ is the basis of all possible monomials in $d$ variables of degree $s$
and the norm of a polynomial function $f = \sum_{k \in \KK} w_k \phi_k$ is $\norm{\bm{w}}_p^p = \sum_{k \in \KK} \abs{w_k}^p$.

\paragraph{Exponential tensor kernel {\normalfont :}}
\begin{equation*}
K(\bm{x}^\prime_1,\dots, \bm{x}^\prime_q) 
= \prod_{j=1}^d e^{x^\prime_{1,j}\cdots x^\prime_{q,j}}\\
= e^{\mathrm{sum}(\bm{x}^\prime_1 \odot \cdots \odot \bm{x}^\prime_q)}.
\end{equation*}
This kernel provides an example of an infinite dimensional model, where, $\KK = \N^d$
and, for every $k \in \N^d$, the $k$-th component of the feature map
is $\phi_k(\bm{x}) = \big(1/\prod_{j=1}^d k_j ! \big)^{1/q} \bm{x}^k$.

\subsection{A dual algorithm}
\label{subsec:dualalgo}

In this section we present the main contribution of this paper which is 
an algorithm for solving the problem
\begin{equation}
\label{eq:gen1primal}
\min_{\bm{w} \in \ell^p(\KK)}\!
\!\gamma  \sum_{i=1}^n \big(y_i - \scal{\Phi(\bm{x}_i)}{ \bm{w}}\big)^2 \!+\! \frac 1 p \norm{\bm{w}}_p^p:=F(\bm{w}),
\end{equation}
where $p=q/(q-1)$ with $q$ an even integer (strictly) grater than $2$, 
$\gamma>0$, $\displaystyle\Phi\colon \XC \to \ell^{q}(\KK)$ is the feature map, 
$\KK$ is a countable set, and
$(\bm{x}_i,y_i)_{1 \leq i \leq n} \in (\XC \times \YC)^n$ is the training set. 
Note that \eqref{eq:gen1primal} reduces to \eqref{eq:ellp} if $\mathcal{X} = \R^d$, $\mathbb{K} = \{1,\dots, d\}$ and $\Phi$ is the 
identity map.
The proposed algorithm is based on the minimization of the dual problem \eqref{eq:dualellp2},
where $K$ is defined as in \eqref{def:tensorkernel2}.
This method has two significant characteristics: first, it
is entirely formulated in terms of the tensor kernel function, therefore 
it can also cope with 
nonparametric (infinite dimensional) tensor kernels, 
e.g, the exponential-tensor kernel; 
second, it provides fast convergence.
From the optimization viewpoint, we observe that
the objective functions in \eqref{eq:gen1primal}  and \eqref{eq:dualellp2} are smooth.
However, none of the two has Lipschitz continuous gradient, since in \eqref{eq:gen1primal} $1<p<2$
and in \eqref{eq:dualellp2} the first term is a convex polynomial of degree $q>2$.
This poses an issue
since most gradient algorithms requires Lipschitz continuous gradient to 
achieve convergence \cite{Beck09,Comb05,Dau04}.
Relaxing this assumption for the more general proximal gradient algorithm 
has been the objective of a number of recent works \cite{Cruz15,Bon15a,sal17a}
that introduce suitable linesearch procedures 
to determine the gradient stepsizes.
In light of these studies, 
we present a dual gradient descent algorithm with a backtracking linesearch 
procedure and we prove that,
by exploiting the strong convexity of the dual objective function and the dual-primal link,
the corresponding primal iterates converge linearly to the solution of \eqref{eq:gen1primal}.

To simplify the exposition we treat here the case $q=4$, that is $p=4/3$.
Since the \emph{Gram tensor} 
$\bm{\mathsf{K}} = (K(\bm{x}_{i_1}, \bm{x}_{i_2}, \bm{x}_{i_3}, \bm{x}_{i_4}))_{ i \in \{1, \dots n\}^4}$ 
is of order $4$, it can be viewed as a 
$n^2\times n^2$ symmetric matrix: using a MATLAB-like notation, we define 
$[\bm{\mathsf{K}}] = \mathrm{reshape}(\bm{\mathsf{K}}, n^2, n^2)$.
Likewise, for a $n\times n$ matrix $\bm{B}$, we set
$[\bm{B}] = \mathrm{reshape}(\bm{B}, n^2, 1)$ for its vectorization.
Then, the dual problem \eqref{eq:dualellp2} can be equivalently written as
\begin{equation}
\label{eq:dualellp3}
\min_{\bm{\alpha} \in \R^d}  
\frac 1 q \scal{[\bm{\alpha}\otimes \bm{\alpha}]}{[\bm{\mathsf{K}}] [\bm{\alpha} \otimes \bm{\alpha}]} 
+ \frac{1}{2 \gamma} \norm{\bm{\alpha}}^2 - \scal{\bm{y}}{\bm{\alpha}}:= \Lambda(\bm{\alpha}).
\end{equation}

The proposed dual algorithm is detailed below.

\begin{algorithm}
\label{algo:main}
Let $\bm{\alpha}_0 \in \R^n$,  $\delta, \theta \in \left]0,1\right[$, and initialize the sequence  
$(\lambda_m)_{m \in \N}$ as the constant value $\bar{\lambda} \in \left]0,\gamma/(2(1-\delta))\right[$. 
Then, for every $m \in \N$, 
\begin{equation}
\label{eq:genaccalgo}
\begin{array}{l}
\begin{array}{l}
\bm{\omega}_m = \mathrm{reshape}([\bm{\mathsf{K}}] 
[\bm{\alpha}_m \otimes \bm{\alpha}_m], n,n) \bm{\alpha}_m\ 
\text{(the gradient of the quartic part of $\Lambda$)}\\[1ex]
\nabla \Lambda(\bm{\alpha}_m) = \bm{\omega}_m - \bm{y}+ \gamma^{-1} \bm{\alpha}_m\\[1ex]
\text{while $\Lambda(\bm{\alpha}_m) - \Lambda(\bm{\alpha}_{m} 
- \lambda_m \nabla \Lambda(\bm{\alpha}_m))< \lambda_m (1-\delta) \norm{\nabla \Lambda(\bm{\alpha}_m)}^2$ do}\\[1ex]
\hspace{-1mm}\vspace{2mm}\left\lfloor
\begin{array}{l}
\lambda_m := \theta \lambda_m \\
\end{array}
\right.\\
\bm{\alpha}_{m+1} = (1 - \lambda_m \gamma^{-1}) \bm{\alpha}_m 
- \lambda_m (\bm{\omega}_m - \bm{y})\\
\end{array}
\end{array}
\end{equation}
\end{algorithm}

\begin{remark}
Algorithm~\ref{algo:main} is given for $q=4$. If $q$ is an even integer greater than $4$,
then the leading term of $\Lambda$ is a polynomial of degree $q$ in the variables 
$\bm{\alpha}=(\alpha_1, \dots, \alpha_m)$, and the formula for
 its gradient $\bm{\omega}_m$ at $\bm{\alpha}_m$, even if possibly more complicated, can be still 
 expressed in term of the Gram tensor $\bm{\mathsf{K}}$.
\end{remark}

Our main technical result is the following theorem studying the convergence of the above algorithm.

\begin{theorem}
\label{thm:main}
Let $(\bm{\alpha}_m)_{m \in \N}$ and $(\lambda_m)_{m \in \N}$  
be generated by Algorithm~\ref{algo:main}. Then
we have $\inf_{m} \lambda_m>0$ and, for every $m \in \N$, setting
$\bm{w}_m = J_q\big(\sum_{i=1}^n \alpha_{m,i}\Phi(\bm{x}_i) \big)$, it holds
\begin{equation*}
\norm{\bm{w}_m - \bar{\bm{w}}}_p^2 \leq \dfrac{ \big[(2^p q)
\big(\Lambda(\bm{\alpha}_0) + (\gamma/2) \norm{\bm{y}}_2^2\big)\big]^{\frac{2-p}{p}}
}{C_p} 
\cdot\bigg(1 - \frac{2}{\gamma} \lambda_m(1 - \delta)\bigg)^m
\big( \Lambda(\bm{\alpha}_0) - \min \Lambda\big),
\end{equation*}
for some constant $C_p>0$, depending only on $p$, which tends to zero as $p\to 1$.
Therefore, $\bm{w}_m$ converges linearly to the solution $\bar{\bm{w}}$ 
of problem \eqref{eq:gen1primal}.
\end{theorem}

\begin{remark}
An output $\bm{\alpha}_m = (\alpha_{m,1},\dots, \alpha_{m,n})$ of Algorithm~\ref{algo:main} provides an estimator $\scal{\bm{w}_m}{\Phi(\cdot)}$,
that can be expressed in terms of the tensor kernel $K$ 
through the equation
\begin{equation*}
\scal{\bm{w}_m}{\Phi(\cdot)} =
\sum_{i_1,\dots, i_{q-1}=1}^n  
K(\bm{x}_{i_1}, \cdots, \bm{x}_{i_{q-1}}, \cdot )
\alpha_{m,i_1} \cdots \alpha_{m,i_{q-1}}.
\end{equation*}
Indeed, this follows from recalling the definition of $\bm{w}_m$ in Theorem~\ref{thm:main}
and by reasoning as in \eqref{eq:20170514a}.
\end{remark}

\begin{remark}
\label{rmk:20171013a}
If $p \in \left]1,2\right[$ is not of the form $p=q/(q-1)$ for some $q$ as in \eqref{q-hypothesis}, 
Theorem~\ref{thm:main} remains valid provided that in Algorithm~\ref{algo:main}
$\bm{\omega}_m$ is computed directly in terms of the feature map $\Phi$ evaluated at the training points.
Clearly, this case is feasible only if the feature map is finite dimensional, 
that is, if the index set $\mathbb{K}$ is finite.
\end{remark}

In the following we discuss the most significant aspects of  this dual approach.

\paragraph{Cost per iteration.} The complexity of 
Algorithm~\ref{algo:main}  is mainly related to the 
computation of the gradient of the quartic form in \eqref{eq:dualellp3}, which, by exploiting the symmetries of 
$\bm{\alpha}_m\otimes \bm{\alpha}_m$ and $\bm{\mathsf{K}}$, 
 costs (approximatively) $n^2(n+1)^2/4$ multiplications. We remark that
 in the infinite dimensional case this algorithm is the only feasible approach
 to solve problem \eqref{eq:gen1primal}.
 However, even in the case $\mathrm{card}(\KK)<+\infty$,
e.g., for the linear or polynomial tensor kernel, the method may be convenient 
if $n \ll \mathrm{card}(\KK)$.
Indeed a standard gradient-type algorithm on \eqref{eq:gen1primal} costs $2 n \mathrm{card}(\KK)$ multiplications
($2 n d$ in case of \eqref{eq:ellp}).
Therefore, Algorithm~\ref{algo:main} is recommended 
if $n(n+1)^2/8 \leq \mathrm{card}(\KK)$, that is 
\begin{equation}
\label{eq:compucost}
n  \leq 2 \big(\mathrm{card}(\KK) \big)^{1/3}.
\end{equation}
We stress that Algorithm~\ref{algo:main} has a cost per iteration that depends only on the size $n$ of the data set, while any primal approach will depend on the size of $\KK$.
For instance, in the case of polynomial kernels of degree $s$, we have
$\mathrm{card}(\KK) = (d+s-1) \cdots d/s! \geq d^s/s!$,
and this implies that the cost of a gradient algorithm on the primal problem 
grows exponentially with $s$. We also remark that building the Gram tensor  $\bm{\mathsf{K}}$
will further require $d\cdot n^2(n+1)^2/4$ multiplications (and $8\cdot n^4/8$ bytes in space).
However, the Gram tensor is computed once for all and in a validation procedure
 for the regularization parameter $\gamma$, it 
does not need to be recomputed every time.

\paragraph{Rate of convergence.}
As mentioned above our dual algorithm has linear convergence rate
and can be applied for infinite dimensional kernels. We next discuss the comparison 
with primal approaches when the kernel is finite dimensional 
($\mathrm{card}(\KK)<+\infty$).
The basic point is that primal approaches
will allow only for sublinear rates. Indeed,
since the objective function in \eqref{eq:gen1primal} is the sum of two convex smooth functions, 
among the various algorithms, appropriate choices are
$(a)$ a pure gradient descent algorithm with linesearch (the gradient being that of $F$) and
$(b)$ a proximal gradient algorithm (possibly accelerated) with the prox of $(1/p)\norm{\cdot}_p^p$. 
However, concerning $(a)$ and according to \cite{Bon15a,sal17a}, the algorithm converges, 
but, since $1<p<2$, the full gradient of $F$ is not even locally Lipschitz continuous, so, 
the gradient stepsizes may get arbitrarily close to zero, and 
ultimately the algorithm may exhibit very slow convergence with no explicit rate.
Besides, regarding $(b)$, 
the primal objective function in \eqref{eq:gen1primal} is only uniformly convex on bounded sets.
Therefore, standard convergence results \cite{Beck09,Cha15,Comb05} 
ensure only convergence of the iterates (without rate) and sublinear convergence rate for the objective values.
On the other hand, regarding Algorithm~\ref{algo:main},  we observe
that the constant $C_p$, in Theorem~\ref{thm:main}, approaches zero
as $p\to 1$, so when $p$ is close to 1 the linear convergence rate for the $\bm{w}_m$'s may degrade.
In the numerical experiments, we confirm the above theoretical behaviors: 
the dual algorithm often converges in a few iterations (of the order of 20),
whereas a direct gradient descent method (with linesearch or of proximal-type) on 
the primal problem may require thousands of iterations to reach 
the same precision. 

\paragraph{Dealing with general convex loss.}
Above, we considered, for the sake of simplicity, the least squares loss. 
However, the proposed dual approach can be generalized to all other convex loss functions 
commonly used in machine learning: the \emph{logistic loss} and the \emph{hinge loss} for classification and the \emph{$L^1$-loss}, and the \emph{Vapnik-$\varepsilon$-insensitive loss}
for regression.
In these cases the dual objective function is composed of the same 
leading polynomial form as in \eqref{eq:dualellp2}, 
which has locally Lipschitz continuous gradient, and of a possibly  
nonsmooth (convex) function, having however a closed-form proximity operator
(see Example~\ref{ex:losses} in the supplementary material).
Therefore, according to \cite{sal17a}, for general convex losses, 
instead of Algorithm~\ref{algo:main} we use 
a proximal gradient algorithm with linesearch achieving linear convergence or 
sublinear convergence 
depending on the fact that the dual objective function is strongly convex or not. 
In this respect we note that we have linear convergence for the logistic loss
and sublinear convergence for the $\varepsilon$-insensitive loss and the hinge loss.
This extension is treated in the next section.

\section{Main elements of the theoretical analysis}
\label{sec:theory}
In this section we further develop the discussion of the previous section
and provide the theoretical grounds for the dual approach to 
$\ell^p$-norm regularized learning problems. 
The emphasis here is on the duality theory rather than on the tensor kernels.
The results are presented for general loss function
and any real parameter $p>1$. 

The most general formulation of our objective is as follows,
\begin{equation}
\label{eq:genprimal}
\min_{\bm{w} \in \ell^p(\KK)} 
\gamma  \sum_{i=1}^n L(y_i, \scal{\Phi(\bm{x}_i)}{ \bm{w}})\! +\! \frac 1 p \norm{\bm{w}}_p^p:=F(\bm{w}),
\end{equation}
where $ p>1,\gamma>0$,
 $\displaystyle\Phi\colon \XC \to \ell^{q}(\KK)$ is the feature map, 
$(\bm{x}_i,y_i)_{1 \leq i \leq n} \in (\XC \times \YC)^n$ is the training set, and $L\colon \YC \times \R \to \R$ is a loss function which is convex in the second variable. We define the linear \emph{feature operator}
\begin{equation}
\label{featmat}
\Phi_n \colon \ell^p(\KK) \to \R^n,\quad  \Phi_n\bm{w}= {\big(\scal{\Phi(\bm{x}_i)}{\bm{w}}\big)}_{1 \leq i \leq n}.
\end{equation}
Then its adjoint is $\Phi_n^* \colon \R^n \to \ell^q(\KK)$, $\Phi_n^* \bm{\alpha}= \sum_{i=1}^n \alpha_i \Phi(\bm{x}_i)$.
Duality is based on the following.

\begin{theorem}
\label{thm:duality}
The dual problem of \eqref{eq:genprimal} is
\begin{equation}
\label{eq:gendual}
\min_{\bm{\alpha} \in \R^n} 
\frac{1}{q}\norm{\Phi_n^* \bm{\alpha}}^{q}_{q}
+ \gamma \sum_{i=1}^n L^*\Big(y_i, - \frac{\alpha_i}{\gamma}\Big) := \Lambda(\bm{\alpha}),
\end{equation}
where $L^*(y_i, \cdot)$ is the Fenchel conjugate of $L(y_i, \cdot)$.
Moreover, $(i)$ the primal problem has a unique solution, the dual problem has solutions and $\min F = - \min \Lambda$ (strong duality holds); and
$(ii)$ the solutions $(\bar{\bm{w}}, \bar{\bm{\alpha}})$ 
of the primal and dual problems are characterized by the following KKT conditions
\begin{equation}
\label{eq:kkt}
\begin{cases}
\bar{\bm{w}} = J_q( \Phi_n^* \bar{\bm{\alpha}}),\\
\forall\, i \in \{1,\dots, n\}\ \ 
- \frac{\alpha_i}{\gamma} \in \partial L(y_i, \scal{\Phi(\bm{x}_i)}{ \bar{\bm{w}}}),
\end{cases}
\end{equation}
where $\partial L(y_i,\cdot)$ is the subdifferential of $L(y_i, \cdot)$.
\end{theorem}

All the losses commonly used in machine learning admit explicit Fenchel conjugates
and we refer to the supplementary material for explicit examples. 
The connection between the primal and dual problem is further
deepened in the following result.

\begin{proposition}
\label{prop:dual2primal}
Let $\bar{\bm{\alpha}} \in \R^n$ be a solution of the dual problem $\eqref{eq:gendual}$
and let $\bar{\bm{w}} = J_q\big( \Phi_n^* \bar{\bm{\alpha}} \big)$ be the solution of the primal problem \eqref{eq:genprimal}.
Let $\bm{\alpha} \in \R^n$ and set $\bm{w} = J_q\big( \Phi_n^* \bm{\alpha} \big)$. Then
\begin{equation}
\label{eq:dualineq}
\Lambda(\bm{\alpha}) - \min \Lambda 
\geq   \dfrac{C_p}{ \big[(2^{p} q)\big( \Lambda(\bm{\alpha}) + \gamma \norm{\bm{\xi}}_1\big) \big]^{(2-p)/p}} \norm{\bm{w} - \bar{\bm{w}}}^2_p,
\end{equation}
where, for every $i=1,\dots, n$, $\xi_i = \inf L^*(y_i, \cdot)$ and
$C_p>0$ is a constant that depends only on $p$.
\end{proposition}

The above proposition ensures that if an algorithm generates a sequence 
$(\bm{\alpha}_m)_{m \in \N}$
that is minimizing for the dual 
problem $\eqref{eq:gendual}$, i.e., $\Lambda(\bm{\alpha}_m) \to \min \Lambda$, 
then the sequence defined by
$\bm{w}_m = J_q(\Phi_n^* \bm{\alpha}_m)$, $m \in \N$,
converges to the solution of the primal problem. 

Now, for the most significant losses $L$ in machine learning (see Example \ref{ex:losses}
in the supplementary material), the dual problem \eqref{eq:gendual} has the following form
\begin{equation}
\label{eq:201705417a}
\min_{\bm{\alpha} \in \R^n} \varphi_1(\bm{\alpha}) +  \varphi_2(\bm{\alpha})=\Lambda(\bm{\alpha}),
\end{equation}
where $\varphi_1\colon \R^n \to \R$ 
is convex and  smooth with locally Lipschitz continuous gradient 
($\varphi_1$ will include the term $(1/q) \norm{\Phi_n^* \alpha}_q^q$)
and $\varphi_2\colon \R^n \to \R\cup\{+\infty\}$ is proper, 
lower semicontinuous, convex, and admitting a closed-form proximity operator. So, 
 the form \eqref{eq:gendual} is amenable by the proximal gradient algorithm with 
linesearch studied in \cite{sal17a}, which, referring to \eqref{eq:201705417a}, takes the 
following form.

\begin{algorithm}
\label{dualalgo}
Let $\delta \in \left]0,1\right[$, $\bar{\lambda}>0$, and let $\theta \in \left]0,1\right[$. 
Let $\bm{\alpha}_0 \in \R^n$ and define, for every $m \in \N$,
\begin{equation}
\bm{\alpha}_{m+1} = \prox_{\lambda_m \varphi_2}
(\bm{\alpha}_m -  \lambda_m \nabla \varphi_1(\bm{\alpha}_m)),
\end{equation}
where $\lambda_m = \bar{\lambda} \theta^{j_m}$ and $j_m$ is taken as
 the minimum of the indexes  $j \in \N$ such that $\hat{\bm{\alpha}}_m(j) 
 := \prox_{\lambda_m \varphi_2} 
(\bm{\alpha}_m -  \bar{\lambda} \theta^j \nabla \varphi_1(\bm{\alpha}_m))$ satisfies
\begin{equation*}
 \varphi_1\big( \hat{\bm{\alpha}}_m(j)\big) -  \varphi_1\big( \bm{\alpha}_m\big)  
 - \scal{ \hat{\bm{\alpha}}_m(j) - \bm{\alpha}_m}{\nabla \varphi_1(\bm{\alpha}_m)} 
\leq \delta/(\bar{\lambda} \theta^j) \norm{ \hat{\bm{\alpha}}_m(j) - \bm{\alpha}_m}^2.
\end{equation*}
\end{algorithm}

\begin{remark}
In contrast to Algorithm~\ref{algo:main}, Algorithm~\ref{dualalgo} provides rather a general algorithm
where $\varphi_1$ and $\varphi_2$ are set depending on the choice of the different losses.
\end{remark}

\begin{remark}
If $p=q/(q-1)$ and $q$ satisfies \eqref{q-hypothesis}, then
the computation of $\nabla \varphi_1(\bm{\alpha})$ in Algorithm~\ref{dualalgo} 
can be performed in term of the Gram tensor $\bm{\mathsf{K}}$
(for instance, if $q=4$ the gradient of the quartic part of $\varphi_1$ is as in the first line of 
Algorithm~\ref{algo:main}). Moreover, if in addition $L$ is the square loss,
then $\Lambda$ is as in \eqref{eq:dualellp3} and one can take $\varphi_1=\Lambda$ and
$\varphi_2=0$; and hence Algorithm~\ref{dualalgo} reduces to Algorithm~\ref{algo:main}.
\end{remark}

The convergence properties of Algorithm~\ref{dualalgo} are given in the following 
theorem, which, as opposed to Theorem~\ref{thm:main}, is valid for general loss and any $p \in \left]1,2\right]$.

\begin{theorem}
\label{thm:dualalgo}
Let $p \in \left]1,2\right]$.
Define $(\bm{\alpha}_m)_{m \in \N}$ and $(\lambda_m)_{m \in \N}$ 
as in Algorithm~\ref{dualalgo}.
Then, $\inf_{m} \lambda_m>0$ and, for every $m \in \N$, setting
$\bm{w}_m = J_q(\Phi_n^* \bm{\alpha}_m)$, it holds
\begin{equation*}
\norm{\bm{w}_m - \bar{\bm{w}}}_p \leq o(1/\sqrt{m}).
\end{equation*}
Moreover, if $\Lambda$ 
is strongly convex (which occurs for the least square loss and the logistic loss),
then $\bm{w}_m$ converges linearly to $\bar{\bm{w}}$.
\end{theorem}

\section{Numerical Experiments}
\label{sec:numerics}

We made experiments on simulated data in order to assess the 
following three points.\footnote{
All the numerical experiments have been performed in 
MATLAB\textsuperscript{\textregistered} environment, on a
MacBook laptop with Intel Core 2 Duo,  2 Ghz and 4 GB of RAM.
}

\begin{table}[t]
  \caption{Convergence rates}
  \label{rate}
  \centering
  \begin{tabular}{lllll}
    \toprule
  &  \multicolumn{4}{c}{ Number of iterations (rel. precision $10^{-8}$)}   \\
    \cmidrule{2-5}
    Algorithm     & $p=4/3$   & $p=5/4$  & $p=1.1$ & $p=1.05$  \\
    \midrule
    dual GD + linesearch & 12(5) & 15(4)  & 63(22)  &  258(55) \\
    primal GD + linesearch     & $>5000$ & $>5000$ & $>5000$  & $>5000$     \\
    primal FISTA     & 1158        & 1542 & ---   & --- \\
    \bottomrule
  \end{tabular}
\end{table}

\begin{figure}[t]
\centering
\begin{tabular}{cc}
\includegraphics[width=0.52\textwidth]{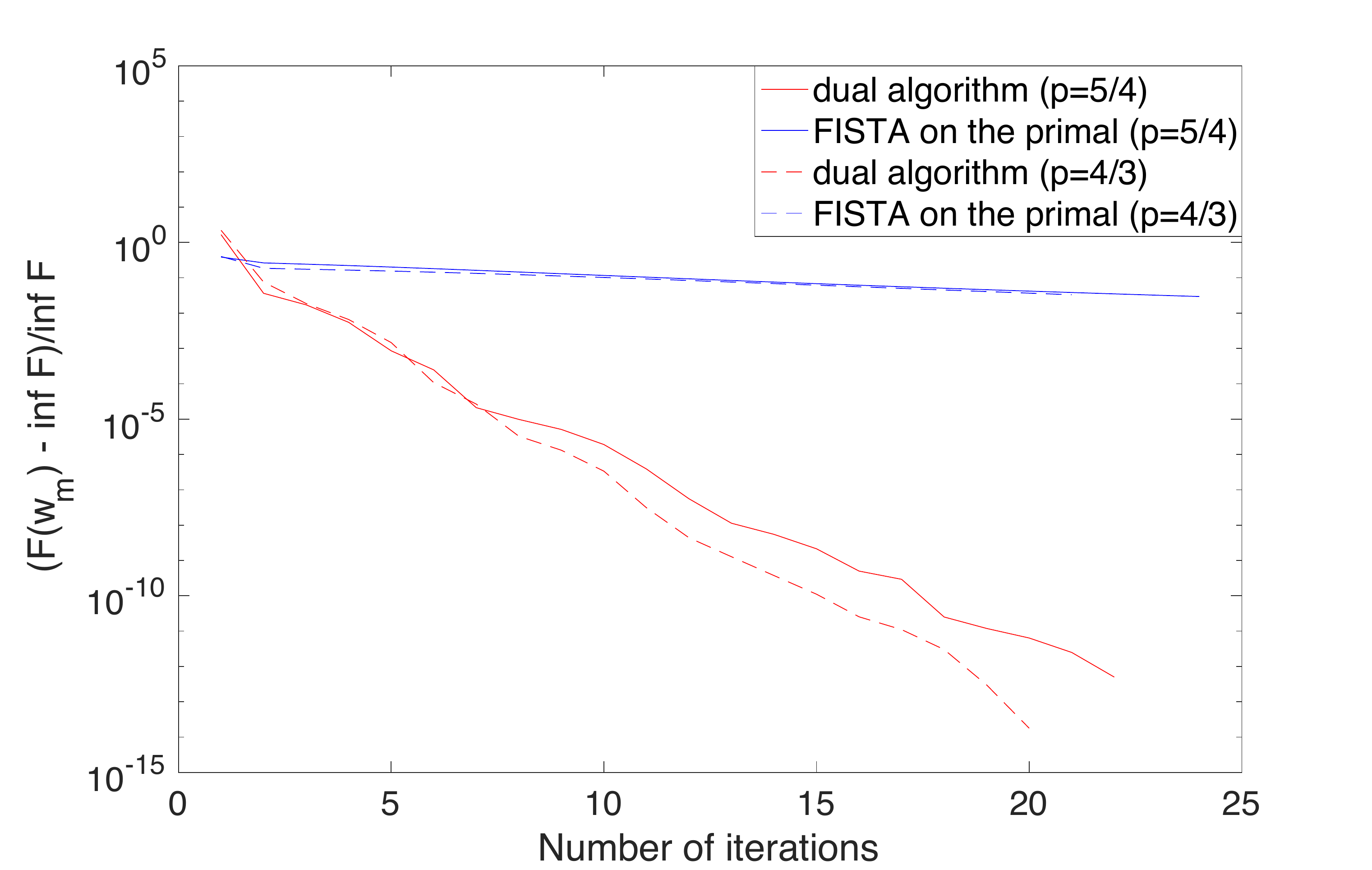}
&\hspace{-6ex}\includegraphics[width=0.52\textwidth]{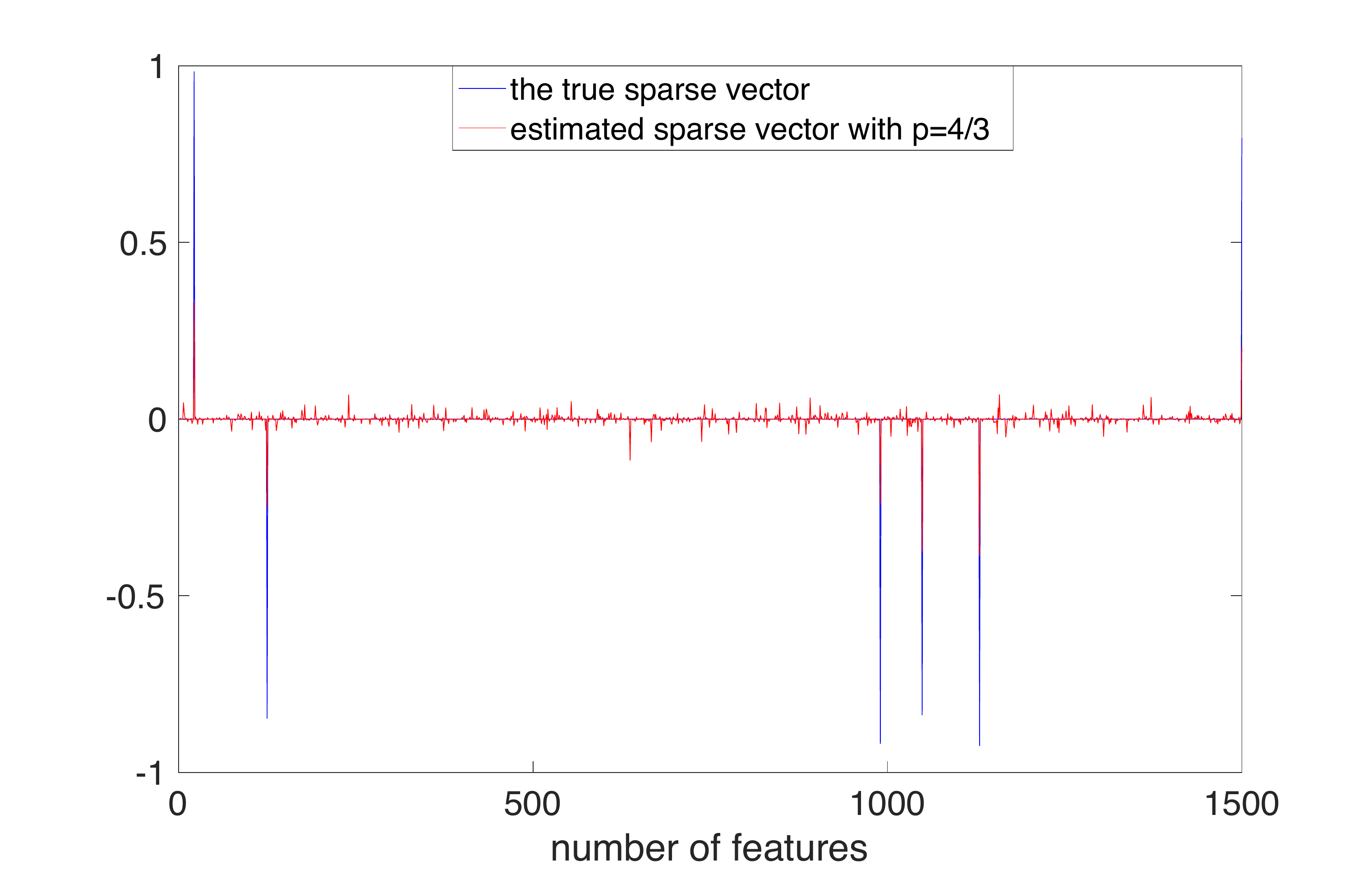}\\
\end{tabular}
\caption{Left. Convergence rates: dual algorithm vs FISTA on the primal. 
Right. True and estimated sparse vectors for a linear tensor kernel: 
$p=4/3$, $n=85$, $d=1500$, and $6$ relevant features.}
\label{fig1}
\end{figure}

\paragraph{Dual vs primal approach (without tensor kernels).}
We considered  problem \eqref{eq:ellp} 
with different choices of $p$ (not necessarily with $q$ even integer).
The purpose is to compare a dual approach against a primal approach per se,
thus without considering the tensor kernel function --- after all 
the dual problem \eqref{eq:dualellp}
is smooth whatever $q$ is.
Algorithm~\ref{algo:main} is therefore modified in such a way that
the gradient of the dual term $(1/q)\norm{\bm{X}^*\bm{\alpha}}_q^q$  is computed
directly as 
$\bm{X}J_q(\bm{X}^*\bm{\alpha})$.\footnote{Note that in this case 
 the cost per iteration is
essentially equal to that of the gradient descent in the primal.}
For the primal approaches we considered two algorithms: $(a)$ the gradient descent method with linesearch
 and $(b)$ the FISTA algorithm \cite{Beck09}, 
 but with $p \in \{4/3, 5/4\}$, since they are the only cases in 
which the proximity operator of $(1/p)\norm{\cdot}_p^p$ 
can be computed explicitly \cite{livre1}.
We generated a matrix $\bm{X}$
according to a normal distribution, a sparse vector $\bm{w}_*$, 
(where the location of the nonzero coefficients was chosen randomly), a 
normal distributed noise vector $\bm{\varepsilon}$, and we defined
\begin{equation*}
\bm{y} = \bm{X} \bm{w}_* + \sigma\bm{\varepsilon},\qquad \sigma = 5\cdot 10^{-2}.
\end{equation*}
We chose $n=200$, $d=10^5$ and $10$ relevant features. The regularization parameter
was set to $\gamma=10$, so to achieve a reconstruction error of the order of the noise.
Table~\ref{rate} and Figure~\ref{fig1}(Left.) clearly show that the dual approach
significantly outperforms  the two primal approaches.\footnote{
The optimal values were found by using the dual algorithm
and checking that the duality gap was $<10^{-14}$.} 

\paragraph{Tensor kernels in the dual approach.}
This experiment considered the case treated in section~\ref{subsec:dualalgo}, that is, $q=4$ 
($p=4/3$),  
with the polynomial tensor kernel of degree $2$, i.e.,
\begin{align*}
K(\bm{x}^\prime_1,\bm{x}^\prime_2,\bm{x}^\prime_3, \bm{x}^\prime_4) 
&= \big(\mathrm{sum}(\bm{x}^\prime_1 \odot \bm{x}^\prime_2\odot \bm{x}^\prime_3 \odot \bm{x}^\prime_4) \big)^2\\
&= \mathrm{sum}(\Phi(\bm{x}^\prime_1) \odot \Phi(\bm{x}^\prime_2)
\odot \Phi(\bm{x}^\prime_3) \odot \Phi(\bm{x}^\prime_4)),
\end{align*}
where $\Phi(\bm{x}) = \big( x_1^2, \dots, x_d^2, \sqrt[4]{2} x_1 x_2, \dots, \sqrt[4]{2} x_1 x_d, 
\sqrt[4]{2} x_2 x_3\dots  \big)$. The dimension of the feature space is $N = d(d+1)/2$.
We generated $\bm{X}$,  $\bm{w}_*$, $\bm{\varepsilon}$ as in the previous case and, according to \eqref{featmat},\footnote{
In this case $\mathrm{card}(\mathbb{K}) = N$, so $\ell^p(\mathbb{K})$ can be identified with $\R^N$
and the linear map $\Phi_n$ can be thought as a $n\times N$ matrix.} 
we defined
\begin{equation*}
\Phi_n = 
\begin{bmatrix}
\Phi(\bm{x}_1)^\top\\
\vdots\\
\Phi(\bm{x}_d)^\top\\
\end{bmatrix} \in \R^{n\times N},
\ \bm{y} = \Phi_n \bm{w}_* + \sigma\bm{\varepsilon}, \  \sigma = 5\cdot 10^{-2}.
\end{equation*}
Then we aimed at solving problem \eqref{eq:ellp} with $\bm{X}$ replaced by $\Phi_n$.
We examined a situation in which the  computational cost per iteration of the dual algorithm
is less than the corresponding primal, measuring the gain in CPU time. 
We set $n=90$, $d=650$ and $6$ relevant features
out of the total of $N = 211575$. With these figures, according to the discussion at the end of
section~\ref{subsec:dualalgo}, computing the gradient through the tensor kernel, as done in 
Algorithm~\ref{algo:main},
surely reduces the cost per iteration. Table~\ref{dual+tensor} shows
the CPU time required by the dual algorithm with and without using the tensor kernel.

\begin{table}[t]
  \caption{The dual algorithm with and without tensor kernels ($p=4/3$).}
  \label{dual+tensor}
  \centering
  \begin{tabular}{lll}
    \toprule
    Algorithm   & CPU time (sec)  & iterations   \\
    \midrule
        build the Gram tensor $\bm{\mathsf{K}}$ & 2.73 & ---\\
    dual GD + linesearch (with $\bm{\mathsf{K}}$) & 2.49 & 29  \\
    dual GD + linesearch (without $\bm{\mathsf{K}}$)  &    9.87 & 28 \\
    \bottomrule
  \end{tabular}
\end{table}

\paragraph{Recovering the relevant features.}
The sparseness properties of an $\ell^p$-regularization method
were mentioned in \cite{Dau04} and later were
studied more carefully in \cite{Kol09}, from a statistical viewpoint.
In contrast to $\ell^1$-regularization, the $\ell^p$-regularization does not generally provide
finite supported vectors, so sparseness here actually means  \emph{approximate sparsity}
in the sense that the insignificant coefficients are shrunk and the relevant ones are 
highlighted.
Our experiments confirm this property of $\ell^p$ regularization.
Indeed in the setting described in the previous scenarios, 
 the solution vector $\bar{\bm{w}}$ always exhibits spikes that 
 corresponds to the non zero coefficients of $\bm{w}_*$.
Depending on the value of $p$, on the size $n$ of the data set, and on
the feature space dimension $N$, this phenomenon may be more or less notable,
but in any case the vector $\bar{\bm{w}}$
either clearly reveals the hidden relevant features (see Figure~\ref{fig1}(Right.)) or 
can be safely thresholded in order to discard most of the non-relevant features,
and reduce the dimensionality of the problem of $1$-$2$ orders of magnitude.

\section{Conclusions}

In this paper we presented a novel and efficient kernel method for $\ell^p$-norm
regularized learning problems.
The method assumes that $p=q/(q-1)$ with $q$ an even integer grater than $2$.
In such case,
we provided an algorithm which is based on the minimization of the dual problem
and can be formulated in terms of a tensor kernel evaluated at the training points,
avoiding the call of the feature map. 
Therefore,  this 
 provides the first viable solution to $\ell^p$-type 
regularization in infinite dimensional spaces.
Moreover,  in finite dimension,
the proposed approach compares favorably to other solutions in the regime
of few sample and large number of variables, and $q$ reasonably low.
For example, our experiments show that if $q=4$, the proposed method is practicable and provides an effective variables
selection method and/or is able to discard most of the irrelevant features. 
We remark that, the complexity of the method depends only on the dataset size and does not depend 
on the dimension of the function space (e.g, the degree of the polynomial kernel).
However, there are scenarios and values of $q$ in which using tensor kernels may be cumbersome
from the computational point of view, but this difficulty is common to other approaches to 
nonparametric sparsity and it is certainly a challenge that requires further study.
Finally, the experiments are meant to provide a proof of concept for the proposed method
and are the starting point for a more systematic empirical study that we defer to a future work.

\bibliographystyle{plain}
\bibliography{publications}

\newpage

\appendix
\section{Appendix}

This section contains proofs and additional details on some of the topics discussed above.

\subsection{Duality in $\ell^p$-regularization}

{\em Proof of Theorem~\ref{thm:duality}.}
Problem \eqref{eq:genprimal} can be written in the form
\begin{equation}
\label{FRduality1}
\min_{w \in \ell^p(\KK)} f(\bm{w}) + g(-\Phi_n \bm{w}),
\end{equation}
where $f(\bm{w}) = (1/p) \norm{\bm{w}}_p^p$, $g(\bm{\beta}) 
= \gamma\sum_{i=1}^n L(y_i, -\beta_i)$,
and $\Phi_n$ is defined as in \eqref{featmat}. The Fenchel-Rockafellar dual  problem of 
\eqref{FRduality1} is \cite{livre1}
\begin{equation}
\min_{\bm{\alpha} \in \R^n} f^*(\Phi_n^* \bm{\alpha}) + g^*(\bm{\alpha})
\end{equation}
and the corresponding KKT optimality conditions are
\begin{equation*}
\bar{\bm{w}} \in \partial f^* (\Phi_n^* \bar{\bm{\alpha}})\qquad\text{and}\qquad
\bar{\bm{\alpha}} \in \partial g(-\Phi_n \bar{\bm{w}}).
\end{equation*}
Now it is easy to see that
\begin{equation*}
(\forall\, \bm{\alpha} \in \R^n)\quad g^*(\bm{\alpha}) 
= \gamma \sum_{i=1}^n L^*\Big(y_i, -\frac{\alpha_i}{\gamma}\Big)
\qquad \text{and}\qquad (\forall\, \bm{u} \in \ell^q(\KK))\quad  f^*(\bm{u}) 
= \frac{1}{q}\norm{\bm{u}}_q^q.
\end{equation*}
Therefore, the dual form $\eqref{eq:gendual}$ follows. Statement $(i)$ comes from the fact that
$g$ is continuous. Statement $(ii)$ follows from the KKT conditions above by noting
that $f^*$ is indeed differentiable and $\nabla f^* = J_q$, and the fact that $g$ is separable.
\qed

We now specialize Theorem~\ref{thm:duality} to distance-based and  margin-based 
losses [3, Definitions~2.24 and 2.32].

\begin{corollary}
Suppose that $L$ is a convex distance-based loss of the form $L(y,t) = \psi(y - t)$
with $\YC=\R$, for some convex function $\psi\colon \R \to \R_+$. 
Then the dual problem $\eqref{eq:gendual}$
becomes
\begin{equation}
\label{eq:dual1}
\min_{\alpha \in \R^n} 
\frac{1}{q}\norm{\Phi_n^* \bm{\alpha}}^{q}_{q} - \bm{y}^\top \bm{\alpha}
+ \gamma \sum_{i=1}^n  \psi^* \Big(\frac{\alpha_i}{\gamma} \Big)
\end{equation}
Suppose that $L$ is a convex margin-based loss of the form $L(y,t) = \psi(y t)$
with $\YC=\{-1,1\}$, for some convex function $\psi\colon \R \to \R_+$. 
Then the dual problem $\eqref{eq:gendual}$
becomes
\begin{equation}
\label{eq:dual2}
\min_{\alpha \in \R^n} 
\frac{1}{q}\norm{\Phi_n^* \bm{\alpha}}^{q}_{q} 
+ \gamma \sum_{i=1}^n  \psi^* \Big(- \frac{y_i \alpha_i}{\gamma} \Big).
\end{equation}

\end{corollary}

The following example shows that 
all the losses commonly used in machine learning admit explicit Fenchel conjugates.

\begin{example}\ 
\label{ex:losses}
\begin{enumerate}[$(i)$]
\item The \emph{least squares loss} is  $L(y,t) = \psi(y - t)$ with $\psi = (1/2) \abs{\cdot}^2$.  
In that case \eqref{eq:dual1} reduces to \eqref{eq:dualellp}, which is
 is strongly convex with modulus $1/\gamma$.
\item The \emph{Vapnik-$\varepsilon$-insensitive loss} for regression is $L(y,t) = \psi(y - t)$ with $\psi = \abs{\cdot}_\varepsilon$. Then, $\psi^* = \varepsilon \abs{\cdot} + \iota_{[-1,1]}$
and the last term in \eqref{eq:dual1} turns out to be $\varepsilon \norm{\bm{\alpha}}_1 
+ \iota_{\gamma[-1,1]^n}(\bm{\alpha})$.
\item The \emph{Huber loss} is the distance-based loss defined by
\begin{equation*}
\psi(r) = 
\begin{cases}
r^2/2 &\text{if } \abs{r} \leq \rho\\
\rho \abs{r} - \rho^2/2 &\text{otherwise}.
\end{cases}
\end{equation*}
Then $\psi^* = \iota_{[-\rho, \rho]} + (1/2) \abs{\cdot}^2$ [1, Example~13.7] and 
the last term in \eqref{eq:dual1}
is $(1/(2\gamma))\norm{\bm{\alpha}}_2^2 
+ \iota_{\rho\gamma[-1, 1]^n}(\bm{\alpha})$.
\item The \emph{logistic loss} for classification is the margin-based loss with $\psi(r) = \log (1+e^{-r})$.
Thus
\begin{equation*}
\psi^*(s) = 
\begin{cases}
(1+s)\log(1+s)-s\log(-s) &\text{if } s \in \left]-1,0\right[\\
0&\text{if } s = -1 \text{ or } s = 0\\
+\infty&\text{otherwise}.
\end{cases}
\end{equation*}
See \cite[Example~13.2(vi)]{livre1}.
It is easy to see that $\psi$ has Lipschitz continuous derivative with constant $1/4$
and hence $\psi^*$ is strongly convex with modulus $4$ \cite{livre1}. Thus,
referring to \eqref{eq:dual1}, we see that
in this case $\dom \Lambda = \prod_{i=1}^n (y_i [0,\gamma])$ and 
$\Lambda$ is differentiable on $\mathrm{int}(\dom \Lambda)$ with locally
Lipschitz continuous gradient. Moreover, since $\lim_{s \to1} \abs{(\psi^*)^\prime(s)}
= \lim_{s \to0} \abs{(\psi^*)^\prime(s)} = +\infty$, we have that 
$\norm{\nabla \Lambda (\bm(\alpha))} = +\infty$ on the boundary of $\dom \Lambda$.
Finally, it follows from 
\eqref{eq:kkt} that $0<y_i \bar{\alpha}_i < \gamma$, for $i=1,\dots, n$.
Note that in this case we can still apply 
Algorithm~\ref{dualalgo} with $\varphi_2=0$ 
(see \cite[Section~4]{sal17a}).
\item The \emph{hinge loss} is the margin-based loss with $\psi(r) = (1-r)_+$. 
We have $\psi^*(s) = s + \iota_{[-1,0]}(s)$. So the second term in \eqref{eq:dual2} 
is $-\sum_{i=1}^n y_i \alpha_i + \iota_{\gamma[0,1]}(y_i \alpha_i)$
\end{enumerate}
We also note that in all cases, for every $i \in \{1,\dots, n\}, \inf L^*(y_i,\cdot)>-\infty$,
which was a condition considered in Proposition~\ref{prop:dual2primal}.
\end{example}

{\em Proof of Proposition~\ref{prop:dual2primal}.}
We use the same notation as in the proof of Theorem~\ref{thm:duality}.
It follows from the definitions of 
$\bm{w}$ and $\bar{\bm{w}}$ and
the Young-Fenchel equalities \cite{livre1} that $f(\bar{\bm{w}}) + f^*(\Phi_n^* \bar{\bm{\alpha}}) 
= \scal{\bar{\bm{w}}}{ \Phi_n^* \bar{\bm{\alpha}}}$
and $f(\bm{w}) + f^*(\Phi_n^* \bm{\alpha}) = \scal{\bm{w}}{\Phi^* \bm{\alpha}}$, and hence
\begin{equation*}
f^*(\Phi_n^* \bm{\alpha}) - f^*(\Phi_n^* \bar{\bm{\alpha}}) = f(\bar{\bm{w}}) - f(\bm{w}) 
+ \scal{\Phi_n^* \alpha}{\bm{w}} - \scal{\Phi_n^* \bar{\bm{\alpha}}}{\bar{\bm{w}}}.
\end{equation*}
Since $-\Phi \bar{\bm{w}} \in \partial g^*(\bar{\bm{\alpha}})$, we have
\begin{equation*}
g^*(\bm{\alpha}) - g^*(\bar{\bm{\alpha}}) 
\geq \scal{-\Phi_n \bar{\bm{w}}}{\bm{\alpha} - \bar{\bm{\alpha}}} 
= \scal{\Phi_n^* \bar{\bm{\alpha}} }{\bar{\bm{w}}}-  \scal{\Phi_n^* \bm{\alpha}}{\bar{\bm{w}}}.
\end{equation*}
Summing the two inequalities above, we get
\begin{align*}
\Lambda(\bm{\alpha}) - \Lambda(\bar{\bm{\alpha}}) 
&\geq f(\bar{\bm{w}}) - f(\bm{w}) -  \scal{ \Phi_n^* \bm{\alpha}}{\bar{\bm{w}} -\bm{ w}}\\
& = f(\bar{\bm{w}}) - f(\bm{w}) - \scal{\nabla f(\bm{w})}{\bar{\bm{w}} - \bm{w}},\\
& = \frac 1 p \norm{\bar{\bm{w}}}_p^p - \frac 1 p \norm{\bm{w}}_p^p - \scal{J_p(\bm{w})}{\bar{\bm{w}} - \bm{w}}.
\end{align*}
Now, since $1<p<2$, it follows from [2, Corollary~2.6.1] that
\begin{equation*}
\frac 1 p \norm{\bar{\bm{w}}}_p^p - \frac 1 p \norm{\bm{w}}_p^p - \scal{J_p(\bm{w})}{\bar{\bm{w}} - \bm{w}}\\
\geq
\dfrac{C_p}{\big(\norm{\bar{\bm{w}}}_p + \norm{\bm{w}}_p \big)^{2 - p}} \norm{\bar{\bm{w}} - \bm{w}}^2_p,
\end{equation*}
for some constant $C_p>0$ that depends only on $p$.
Therefore, by the definition of the duality map, 
\begin{equation*}
\norm{\bm{w}}_p = \norm{J_q(\Phi_n^* \bm{\alpha})}_p 
= \big(\norm{\Phi_n^* \bm{\alpha}}^q_q\big)^{1/p}
\leq q^{1/p} \big( \Lambda(\bm{\alpha}) + \gamma\norm{\bm{\xi}}_1\big)^{1/p},
\end{equation*}
where $\xi_i = \inf L^*(y_i, \cdot)$;
and similarly for $\norm{\bar{\bm{w}}}_p$. Then the statement follows.
\qed

{\em Proof of Theorem~\ref{thm:main}.} Since for the least squares loss we have
$\xi_i= - (1/2) y_i^2$, it follows from Proposition~\ref{prop:dual2primal} that 
for every $m \in \N$,
\begin{equation*}
\norm{\bm{w}_m - \bar{\bm{w}}}_p^2 \leq \frac{\big[ (2^p q) 
\big(\Lambda(\bm{\alpha}_m) 
+ (\gamma/2) \norm{\bm{y}}_2^2 \big)\big]^{(2-p)/p}}{C_p}
\big( \Lambda(\bm{\alpha}_m) - \min \Lambda \big).
\end{equation*}
Now it remains to prove that, $\inf_{m} \lambda_m> 0$ and that
\begin{equation}
\label{eq:20170517b}
(\forall\, m \in \N)\qquad \Lambda(\bm{\alpha}_{m+1}) - \min \Lambda 
\leq \big(1 - (2/\gamma)  \lambda_m (1-\delta)\big) 
\big( \Lambda(\bm{\alpha}_{m}) - \min \Lambda  \big).
\end{equation}
First of all, since $q>2$, 
the gradient of $\Lambda$ is Lipschitz continuous on bounded sets.
Therefore, Proposition~3.15 in \cite{sal17a} yields that $\inf_{m} \lambda_m> 0$.
Now, because of the linesearch rule we have that
\begin{equation*}
\Lambda(\bm{\alpha}_{m+1}) 
\leq \Lambda(\bm{\alpha}_{m}) - \lambda_m (1 - \delta) \norm{\nabla \Lambda(\bm{\alpha}_m)}_2^2
\end{equation*}
and, since $\Lambda$ is strongly convex with modulus $1/\gamma$, we have
\begin{equation*}
\Lambda(\bm{\alpha}_m) - \Lambda(\bar{\bm{\alpha}}) 
\leq \frac{\gamma}{2} \norm{\nabla \Lambda(\bm{\alpha}_m)}_2^2.
\end{equation*}
All together the two inequalities above gives
\begin{equation*}
\Lambda(\bm{\alpha}_{m+1}) 
\leq \Lambda(\bm{\alpha}_{m}) - (2/\gamma) \lambda_m (1 - \delta) 
\big(\Lambda(\bm{\alpha}_m) - \Lambda(\bar{\bm{\alpha}}) \big).
\end{equation*}
Adding $\Lambda(\bar{\bm{\alpha}})$ to both sides,
\eqref{eq:20170517b} follows and hence the statement.
\qed

\subsection{The function Banach space associated to a tensor kernel}
\label{appendixA2}

In this section we make explicit the space associated to tensor kernels.
We assume that $\mathrm{span}(\Phi(\R^d))$ is dense in $\ell^q(\KK)$ --
which is equivalent to requiring that the functions $(\phi_k)_{k \in \KK}$ are $\ell^p$ point-wise independent. Then, we can associate 
to the feature map $\Phi$ the Banach function space \cite{Zhan09}
\begin{equation}
\label{eq:rkbs}
\mathcal{B} = \big\{ \scal{\bm{w}}{\Phi(\cdot)} \,\big\vert\, \bm{w} \in \ell^p(\KK)\big\},
\qquad \norm{\scal{\bm{w}}{\Phi(\cdot)}}_{\mathcal{B}} = \norm{\bm{w}}_p.
\end{equation}
Note that if $\bm{\alpha} \in \R^n$, $\bm{x}_1, \dots, \bm{x}_n \in \R^d$, and we set 
$\bm{w} = J_q \big(\sum_{i=1}^n \alpha_i \Phi(\bm{x}_i) \big)$, then, as in \eqref{eq:20170511b},
we have 
\begin{equation}
\label{eq:20170512a}
\scal{\bm{w}}{\Phi(\cdot)} = \sum_{i_1,\dots, i_{q-1}=1}^n  
K(\bm{x}_{i_1}, \cdots, \bm{x}_{i_{q-1}}, \cdot )
\alpha_{i_1} \cdots \alpha_{i_{q-1}},
\end{equation}
and
\begin{equation}
\label{eq:20170519a}
\norm{\scal{\bm{w}}{\Phi(\cdot)}}_{\mathcal{B}} = \bigg( \sum_{i_1, \dots, i_q = 1}^n 
K(\bm{x}_1, \dots, \bm{x}_q) \alpha_{i_1}\cdots \alpha_{i_q} \bigg)^{1/p},
\end{equation}
and the functions \eqref{eq:20170512a} are dense in $\mathcal{B}$.
Moreover, setting $\Phi^* = J_q \circ \Phi \colon \R^d \to \ell^p(\KK)$,
if $\mathrm{span}(\Phi^*(\R^d))$ is also dense in $\ell^p(\KK)$, then its associated
function Banach space $\mathcal{B}^*$ (defined similarly to \eqref{eq:rkbs}) 
is the topological dual of $\mathcal{B}$
and the following reproducing property holds
\begin{equation*}
K_{\bm{x}}\colon \bm{x}^\prime \to K(\bm{x}^\prime,\dots, \bm{x}^\prime, \bm{x}) \in \mathcal{B}^*,\quad\text{and}
\quad \scal{f}{K_{\bm{x}}} = f(\bm{x}).
\end{equation*}
For the case of infinite dimensional power series tensor kernels, which includes the exponential
tensor kernels considered here, the density assumptions on $\mathrm{span}(\Phi(\R^d))$
and $\mathrm{span}(\Phi^*(\R^d))$ holds, hence the corresponding Banach space can be 
described through the equations \eqref{eq:20170512a} and \eqref{eq:20170519a}.
\subsection{The dual algorithm for general loss function and any $p \in \left]1,2\right[$}

{\em Proof of Theorem~\ref{thm:dualalgo}.}
Since $\varphi_1$ is smooth with a locally Lipschitz continuous gradient
we can apply Theorem~3.2 and Proposition~3.5 in \cite{sal17a} and get 
$\inf_m \lambda_m>0$,
$\bm{\alpha}_m \to \bar{\bm{\alpha}}$ and
$\Lambda(\bm{\alpha}_m) - \Lambda(\bar{\bm{\alpha}})= o(1/m)$. Then, by Proposition~\ref{prop:dual2primal}, 
we have $\norm{\bm{w}_m - \bar{\bm{w}}}_p \leq o(1/\sqrt{m})$.
Now suppose that $\Lambda$ is $\mu$-strongly convex. We will rely on
Proposition~2 in \cite{Bre08}.
Then, strong convexity of $\Lambda$ yields
\begin{equation*}
\frac \mu 2 \norm{\bm{\alpha}_m - \bar{\bm{\alpha}}}^2 
\leq \Lambda(\bm{\alpha}_m) - \Lambda(\bar{\bm{\alpha}})
\end{equation*}
for some constant $\mu>0$. So equation (3.8) in Proposition~2 in \cite{Bre08} holds.
Moreover, defining
\begin{equation*}
- D_{\lambda_m} (\bm{\alpha}_m)  := 
\varphi_2 (\bm{\alpha}_{m+1}) \big) - \varphi_2 (\bm{\alpha}_{m}) \big)
+ \scal{\bm{\alpha}_{m+1} - \bm{\alpha}_m}{\nabla \varphi_1(\bm{\alpha}_m)},
\end{equation*}
by the definition of $\lambda_m$, and Proposition~3.8 and Proposition~3.9 in \cite{sal17a},
we have
\begin{equation}
\label{eq:20170519a}
\frac{\norm{\bm{\alpha}_{m+1} - \bm{\alpha}_m}^2}{\lambda_m} \leq D_{\lambda_m} (\bm{\alpha}_m)\quad
\text{and}\quad
\Lambda(\bm{\alpha}_{m+1}) - \Lambda(\bm{\alpha}_{m})  
\leq - (1-\delta) D_{\lambda_m} (\bm{\alpha}_m).
\end{equation}
Then, since $\inf_{m} \lambda_m>0$ we can proceed as in the proof of Proposition~2 in \cite{Bre08}
and prove that $\Lambda(\bm{\alpha}_m)$ converge linearly to $\Lambda(\bar{\bm{\alpha}})$.
Finally, using Proposition~\ref{prop:dual2primal} the linear convergence of $\bm{w}_m$  follows.
Note that Example~\ref{ex:losses} shows that if $L$ is the least square loss or
the logistic loss, then $\Lambda$ is strongly convex.
\qed

\end{document}